\documentclass{article} % For LaTeX2e
\usepackage{arxiv_mod,times}

\usepackage{amsmath,amsfonts,bm}

\def\eqref#1{equation~\ref{#1}}
\def\1{\bm{1}}

\DeclareMathAlphabet{\mathsfit}{\encodingdefault}{\sfdefault}{m}{sl}
\SetMathAlphabet{\mathsfit}{bold}{\encodingdefault}{\sfdefault}{bx}{n}

\usepackage{hyperref}
\usepackage{url}
\usepackage{graphicx}
\usepackage{algorithm}
\usepackage{algpseudocode}
\usepackage{amsmath,amssymb}

\newcommand{\Normalize}{\textproc{Normalize}}

\title{Tracing Distinguishability through \\Transformer Processing with \\Stochastic LayerNorm}

\author{
Kieran Murphy \\
New Jersey Institute of Technology \\
\texttt{kieran.murphy@njit.edu}
}

\iclrfinalcopy
\begin{document}

\maketitle

\begin{abstract}

Representational similarity is foundational to analyses of deep networks, yet distances between point-valued representations are not intrinsically tied to downstream function: nearby states may produce different behaviors, while distant states may behave similarly.
We instead give representations volume, turning similarity into statistical distinguishability.
Overlapping stochastic representations necessarily induce overlapping downstream distributions, grounding latent comparison in model function and bringing it under information-theoretic tools such as the data-processing inequality.
We realize this idea in pretrained transformers through a light-touch modification to LayerNorm: at each residual-stream read, we normalize the state, add isotropic Gaussian noise, and renormalize.
During distillation fine-tuning, one learned allocation parameter per residual-stream read distributes a fixed global rate budget across the processing stack. 
The resulting model can be viewed as transformer blocks reading the residual stream with learned finite precision under a shared global rate budget.
Using the Bhattacharyya coefficient, we trace which counterfactual distinctions are preserved through MLP blocks or selectively exposed to the query, key, and value computations of individual attention heads.
Experiments on ViT-S and GPT-2 small reveal the depthwise propagation of continuous visual perturbations and head-specific sensitivity to token distinctions aligned with known attention motifs.
These results establish distinguishability as a functionally grounded lens on transformer computation that complements existing interpretability approaches.

\end{abstract}

\section{Introduction}

Deep networks are ordinarily trained on point-valued representations: each example constrains the computation at a measure-zero set of internal states, while behavior in the surrounding representational volume is left implicit.
Many interpretability methods impose geometric notions of similarity on point-valued representations -- for example, through pairwise similarity measures~\citep{kornblith2019similarity} or the reconstruction distortion used to fit sparse surrogate models~\citep{huben2024sparse,bricken2023monosemanticity} -- even though this geometry is not directly grounded by the model's functional objective.
The disconnect between geometry and functionality when assessing point-based representations has been increasingly recognized in recent years~\citep{ding2021grounding,cui2022deconfounded,davari2023ckareliability}.

When point-valued representations are replaced by stochastic ones, they acquire volume: behavior over neighborhoods of internal states becomes part of the training objective, coupling representational geometry to model functionality.
The resulting stochastic representations define explicit channels through which quantities such as information transmission and pairwise distinguishability can be measured.
Without a constraint on representation scale, however, a model can trivially overcome stochasticity by increasing the separation of its representations, recovering an effectively deterministic channel.
Prior approaches therefore impose explicit capacity constraints, often through learned encoder-decoder bottlenecks~\citep{vae,alemiVIB2016,vit_ib}. 
Doing so at many internal locations requires substantial architectural modification.

Here, we propose a light-touch modification to transformer architectures that exploits the intrinsic scale constraint imposed by LayerNorm~\citep{ba2016layernorm}.
Additive noise is applied to every LayerNorm in a transformer, before the learned affine scale and shift, with one optimized magnitude per location such that only $\sim 2 L$ additional scalar parameters are introduced.
In pre-norm transformers, each modification can be interpreted as a noisy read of the residual stream~\citep{elhage2021mathematical}: every attention and MLP block pays for measurement precision when accessing the representation carried by the residual stream.

\begin{figure*}[t]
\vskip 0.2in
\begin{center}
\centerline{\includegraphics[width=\textwidth]{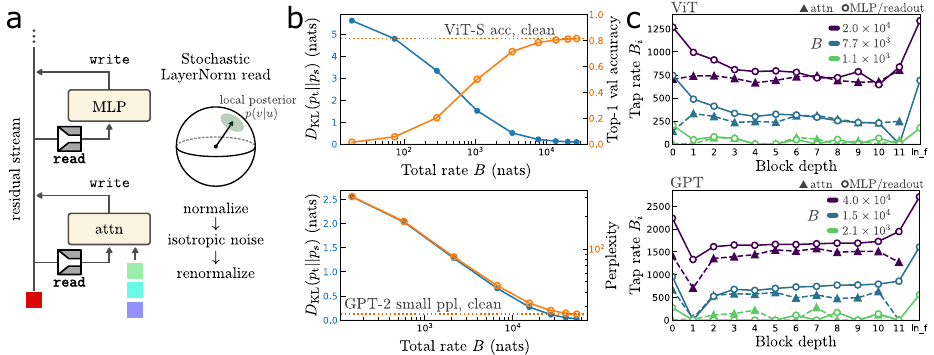}}
\caption{
\textbf{Injecting stochasticity into residual stream reads.}
\textbf{(a)} We add noise to every location in the network where the residual stream is read.
The noise is injected inside the LayerNorm operation, which maps representations to the surface of a hypersphere.
\textbf{(b)} For fine-tuned ViT-S (top) and GPT-2 small (bottom), the imposed read rate $B$ sets the distillation loss relative to the base (deterministic) model, $D_\text{KL}(\text{teacher}||\text{student})$, as well as the modality-specific validation performance.
\textbf{(c)} Per-tap rates provide a macroscopic view of prioritized information protection. 
For each attention block and MLP block, and for the final read before the unembed operation, a single $\sigma$ is learned during optimization while fine-tuning all model weights.}
\label{fig:fig1}
\end{center}
\vskip -0.2in
\end{figure*}

Prior bottleneck approaches interpret useful computation upward from a low-information limit~\citep{vit_ib}.
Here, we begin from a pretrained deterministic solution and move toward lower-fidelity internal channels, jointly adapting the model so that stochastic neighborhoods become part of the learned computation.
Because the model weights are optimized jointly with the stochastic channels, the resulting neighborhoods are not merely tolerated by the pretrained computation; they are incorporated into the computation itself.

In experiments on ViT-S and GPT2-small, we use the fine-tuned models as approximations to the pre-trained base that allow information-theoretic tools to be applied natively.
In the vision modality, we measure the models' ability to resolve different types of image augmentation, of varying magnitudes, across depths and bitrates.
In natural language, we design minimal perturbations and measure the propagation of distinguishability in subsequent residual streams.
We measure the degree of perturbation distinguishability accessed by each attention head's query, key, and value projections.
Interestingly, the heads with the most sensitive key projections include several with high induction scores, as well as heads whose selectivity is weak or absent in canonical attention-pattern statistics.
Monitoring perturbation distinguishability at every residual stream read thus grants a foothold into information processing by transformers that is grounded in downstream functionality, complementing existing methodology founded upon the geometry of point-based representations in ambient space.

\section{Method}

A transformer processes input elements ${x_i}$ through distinct residual streams, whose representations ${u_i^l}$ are updated by successive attention and multilayer perceptron (MLP) sublayers. Attention transmits information between streams, whereas the MLP transforms each stream independently. In a pre-norm transformer, every sublayer begins by \textit{reading} its residual stream through LayerNorm. We introduce stochasticity at these reads.

LayerNorm acts as $\operatorname{LN}(u)=\gamma\odot n_\epsilon(u)+\beta$, where $n_\epsilon(u)=(u-\langle u\rangle)/\sqrt{\operatorname{Var}(u)+\epsilon}$ and $\gamma$ and $\beta$ are learned affine parameters~\citep{ba2016layernorm}. Because $\epsilon$ serves only as a numerical safeguard, we set it to zero in the geometric analysis. The normalized state then has component mean zero and component variance one, implying a Euclidean norm of $\sqrt{d}$. LayerNorm therefore maps representations to the $(d-2)$-sphere
$\mathcal{M}_d={z\in\mathbb{R}^d:\langle z\rangle=0,\ \lVert z\rVert_2=\sqrt{d}}$.

We describe the modification at a single tap, suppressing stream and tap indices except on the tap-dependent noise scale $\sigma_t$. We normalize the residual-stream state, add isotropic Gaussian noise, and renormalize:

$$
    z=n_\epsilon(u),
    \qquad
    v=n_\epsilon(z+\eta),
    \qquad
    \eta\sim\mathcal{N}(0,\sigma_t^2I).
$$

The usual LayerNorm affine transformation is then applied to $v$ before it is passed to the attention or MLP sublayer. The construction is applied identically at every residual read, with independent noise draws and a separate noise scale for each tap.

For a particular input and realization of all upstream noise, $u$ and $z$ are individual points. Let $U$ denote the raw residual-stream state when viewed as a random variable over inputs and upstream noise, let $Z=n_\epsilon(U)$ be its normalized form, and let $V$ be the output of the local stochastic channel. By the data-processing inequality, no subsequent computation using this read can contain more information about $U$, through this read, than is available in $V$. We quantify this fundamental local limitation using Shannon mutual information, $I(U;V)$~\citep{cover1991elements}. Because $Z$ is a deterministic function of $U$ and the channel depends on $U$ only through $Z$, $I(U;V)=I(Z;V)$.

The maximum-entropy distribution on $\mathcal{M}_d$, relative to its natural surface measure, is the uniform distribution. Letting $A_d$ denote the surface area of $\mathcal{M}_d$, its entropy is $\log A_d$. The transmitted information can therefore be upper-bounded as

$$
    I(U;V)
    =
    h(V)-h(V\mid U)
    \leq
    \log A_d-h(V\mid U)
    \equiv R_d(\sigma_t).
$$

Rotational symmetry makes the conditional entropy, and hence the rate bound $R_d(\sigma_t)$, a function only of the noise scale and representation dimension. The gap in the bound is

$$
    R_d(\sigma_t)-I(U;V)
    =
    D_{\mathrm{KL}}
    \left(
        p_V\,\middle\|\,
        \operatorname{Unif}(\mathcal{M}_d)
    \right),
$$

where $p_V$ is the marginal distribution of noisy reads induced by the distribution of $U$ and the local noise. Thus, the bound is tight when the measured states are marginally uniform over the LayerNorm sphere.

A global hyperparameter $B$ fixes the total rate bound across all taps. 
Per-tap rates are learned subject to this fixed total, with allocation logits ${a_t}$ defining $B_t=B\operatorname{softmax}(a)_t$. 
For a given representation dimension, we precompute the monotonic relationship between $\sigma$ and $R_d(\sigma)$ and use this lookup table to convert each allocated rate into its corresponding noise scale. 
Exact zero rate is reached only in the limit $\sigma\rightarrow\infty$; the finite truncation of the lookup table and its corresponding minimum realizable rate are detailed in Appendix~\ref{appx:implementation}.

We install these stochastic reads in a pretrained model and fine-tune the resulting student by distillation from the original deterministic model. Specifically, we minimize

$$
    \mathcal{L}_{\mathrm{distill}}
    =
    \mathbb{E}_{x}
    \left[
        D_{\mathrm{KL}}
        \left(
            p_{\mathrm{teacher}}(\,\cdot\mid x)
            \,\middle\|\,
            p_{\mathrm{student}}(\,\cdot\mid x)
        \right)
    \right].
$$

Unlike information-restriction methods that impose a soft rate penalty~\citep{tishbyIB2000,vit_ib}, our objective contains no rate-loss term. The total rate upper bound is instead enforced directly through the fixed allocation constraint and the calibrated conversion from rates to noise scales.

The modification can be applied to any transformer using LayerNorm. In pre-norm architectures it naturally describes rate-limited reads from the residual stream; when applied after a sublayer in a post-norm architecture, it instead describes rate-limited writes to the residual stream.

\subsection{Tracing distinguishability}

After optimization, the raw residual-stream state remains point-valued, but it affects the downstream sublayer only through a rate-limited stochastic read. The functionally relevant object at a tap is therefore not the point alone, but the conditional distribution of states that the tap can observe. We refer to this distribution as the tap's \textit{local posterior}.

The implemented channel---isotropic Gaussian perturbation followed by renormalization---induces a projected-normal distribution on the LayerNorm sphere. For tractable evaluation, we approximate this distribution with a von Mises--Fisher (vMF) distribution~\citep{davidson2018hyperspherical}. We rescale the LayerNorm sphere to unit radius for this approximation and omit the rescaling from the notation below. Its effective ambient dimension is $p=d-1$, since mean-centering removes one degree of freedom. For a normalized residual state with mean direction $\mu(u)$, the local posterior is approximated as

$$
    p(v\mid u)
    \approx
    C_p(\kappa)\exp\!\left(\kappa\,\mu(u)^\top v\right),
$$

where $C_p(\kappa)$ is the vMF normalizing constant and $\kappa$ controls concentration~\citep{mardia2009directional}. The correspondence between the Gaussian noise scale $\sigma$ and vMF concentration $\kappa$ is calibrated numerically for the relevant dimension and stored in a lookup table. We evaluate the quality of this approximation in Appendix~\ref{appx:gaussian_vmf_calibration}.

Given two raw residual-stream states $u_i$ and $u_j$ at the same tap, we quantify the overlap between their local posteriors using the Bhattacharyya coefficient~\citep{bhattacharyya1943measure}

$$
    \operatorname{BC}(p_i,p_j)
    =
    \int_{\mathcal{S}^{p-1}}
    \sqrt{p(v\mid u_i)\,p(v\mid u_j)}\,dv.
$$

The coefficient ranges from zero for disjoint distributions to one for identical distributions; thus, lower BC indicates greater distinguishability. 
For vMF distributions with a common concentration $\kappa$, integrating the geometric mean of their densities yields the closed form

$$
    \operatorname{BC}(p_i,p_j)
    =
    \frac{C_p(\kappa)}
    {C_p\!\left(
        \kappa\sqrt{\frac{1+\mu_i^\top\mu_j}{2}}
    \right)},
$$

where $\mu_i=\mu(u_i)$ and $\mu_j=\mu(u_j)$.

Processing cannot increase distinguishability. In particular, the squared Hellinger distance, $H^2(P,Q)=1-\operatorname{BC}(P,Q)$, is an $f$-divergence and satisfies the data-processing inequality~\citep{ali1966general,csiszar1967information}. Equivalently, applying a deterministic function or stochastic channel can only increase BC. 
For an MLP sublayer, which transforms each stream independently, the BC between two output distributions must therefore be at least as large as the BC between their reads. 
An attention sublayer instead operates jointly on multiple streams: its output distinguishability is bounded by that of the joint distribution of all visible reads, but not necessarily by the marginal distinguishability of any single read.

\subsubsection{Q, K, V distinguishability}

The linear query, key, and value projections allow us to trace which distinctions in a residual-stream read remain available to each attention head. Under the unit-radius convention adopted above, let $V\sim p(v\mid u)$ denote the local posterior associated with raw residual state $u$. For a query, key, or value matrix $A$, the corresponding head component observes $Y_A=A(\sqrt{d}\,\gamma\odot V+\beta)$. We denote its induced distribution by $p_A(y\mid u)$ and define the projected overlap between two residual states as
\[
    \operatorname{BC}_A(u_i,u_j)
    =
    \operatorname{BC}
    \left(
        p_A(\,\cdot\mid u_i),
        p_A(\,\cdot\mid u_j)
    \right).
\]
Because the projection is generally noninvertible, the data-processing inequality gives $\operatorname{BC}_A(u_i,u_j)\geq\operatorname{BC}(p(\,\cdot\mid u_i),p(\,\cdot\mid u_j))$: a head component may discard distinctions present in its residual-stream read, but cannot create new ones.

To evaluate $\operatorname{BC}_A$, we express $V$ in orthonormal coordinates $X\in\mathcal{S}^{p-1}$ for the mean-zero LayerNorm subspace and let $B_A$ denote the resulting linear map, including the LayerNorm scale $\gamma$. The polar decomposition $B_A=S_AP_A$ separates an invertible output transformation $S_A$ from a row-orthonormal projection $P_A$. Since BC is invariant under $S_A$ and the common output shift, it is sufficient to consider $W_A=P_AX$, whose support is the unit ball. Its pushforward density $g_i^A(w)$ is available in closed form, as derived in Appendix~\ref{appx:qkv_pushforward}. We evaluate the remaining integral by sampling from $r_A(w)=\tfrac{1}{2}[g_i^A(w)+g_j^A(w)]$:
\[
    \widehat{\operatorname{BC}}_A(u_i,u_j)
    =
    \frac{1}{N}
    \sum_{n=1}^{N}
    \frac{
        2\sqrt{g_i^A(w_n)g_j^A(w_n)}
    }{
        g_i^A(w_n)+g_j^A(w_n)
    },
    \qquad
    w_n\sim r_A.
\]
Each summand lies in $[0,1]$, providing a bounded Monte Carlo estimator of the overlap retained by the query, key, or value projection.
\subsubsection{Common random numbers}

At a given target tap, upstream stochasticity makes the effective posterior a mixture of local posteriors. Let $\epsilon_{1:L}$ denote the Gaussian noise draws at the $L$ upstream taps, and let $p(v\mid x,\epsilon_{1:L})$ denote the local posterior conditioned on a complete upstream noise sequence. The effective posterior for input $x$ is
\[
    \bar{p}(v\mid x)
    =
    \mathbb{E}_{\epsilon_{1:L}}
    \left[
        p(v\mid x,\epsilon_{1:L})
    \right].
\]

When comparing two inputs, we couple their evaluations using common random numbers (CRN), a standard variance-reduction technique for paired stochastic simulations~\citep{glasserman1992some}. Specifically, we apply the same sequence of Gaussian noise vectors to both inputs, so that variation between the resulting trajectories is driven primarily by the input difference rather than unrelated noise draws. We average the resulting BC over multiple independently sampled CRN sequences. By the joint concavity of the Bhattacharyya coefficient,
\[
    \operatorname{BC}
    \left(
        \bar{p}(\,\cdot\mid x_1),
        \bar{p}(\,\cdot\mid x_2)
    \right)
    \geq
    \mathbb{E}_{\epsilon_{1:L}}
    \left[
        \operatorname{BC}
        \left(
            p(\,\cdot\mid x_1,\epsilon_{1:L}),
            p(\,\cdot\mid x_2,\epsilon_{1:L})
        \right)
    \right].
\]
Thus, the fully coupled CRN estimate is a lower bound on the BC between the effective posteriors. Joint concavity follows directly by applying the Cauchy--Schwarz inequality to the mixed densities.

One can interpolate between these endpoints by sharing only a prefix of the upstream noise sequence. 
For $0\leq k\leq L$, define the partially marginalized posterior
\[
    p^{(k)}(v\mid x,\epsilon_{1:k})
    =
    \mathbb{E}_{\epsilon_{k+1:L}}
    \left[
        p(v\mid x,\epsilon_{1:L})
    \right],
\]
and the corresponding overlap
\[
    \operatorname{BC}_k
    =
    \mathbb{E}_{\epsilon_{1:k}}
    \left[
        \operatorname{BC}
        \left(
            p^{(k)}(\,\cdot\mid x_1,\epsilon_{1:k}),
            p^{(k)}(\,\cdot\mid x_2,\epsilon_{1:k})
        \right)
    \right].
\]
Operationally, the first $k$ noise draws are matched across inputs, while the remaining draws are marginalized separately within each posterior. At $k=0$, no noise is conditioned upon and $\operatorname{BC}_0$ is the BC between the full effective posteriors. At $k=L$, the complete noise sequence is shared and $\operatorname{BC}_L$ is the fully coupled CRN estimate. Repeated application of joint concavity gives
\[
    \operatorname{BC}_0
    \geq
    \operatorname{BC}_1
    \geq
    \cdots
    \geq
    \operatorname{BC}_L.
\]
This sequence traces the transition from effective-posterior overlap to the overlap between locally conditioned posteriors.

\section{Experiments}
For both vision and natural language, optimization consisted of fine-tuning a point-based (deterministic) model with solely a distillation loss, $\mathcal{L}=\mathbb{E}_{x \sim p(x)}[D_\text{KL}(p_\text{teacher}(y|x)||p_\text{student}(y|x))]$, with $p(x)$ corresponding to the training data.
The noise level was ramped from $\sigma_g=0.05$ (the uniform-equivalent per-tap noise level defining the budget) to its target value during the first 6 of 25 epochs, equivalently decreasing the rate budget from its near-deterministic initial value.
For vision, we finetuned ViT-S~\citep{dosovitskiy2021} from \texttt{timm} with \texttt{vit\_small\_patch16\_224} weights on ImageNet-1k images~\citep{deng2009imagenet}; for language we used Hugging Face's GPT-2 small~\citep{radford2019language} with \texttt{gpt2} weights and the OpenWebText (OWT) dataset~\citep{Gokaslan2019OpenWeb}.
More specifics can be found in Appx.~\ref{appx:implementation}.

Both vision and language models exhibited a knee near $B\sim10^4$ nats of total read-rate: above it, behavior remained close to the base model, whereas below it, distillation loss and validation performance steadily degraded (Fig.~\ref{fig:fig1}b). 
This behaviorally anchored transition motivates viewing the rate sweep as a probe of the base model’s operating resolution. 
The optimized per-tap allocation provides a macroscopic view of information processing inside the model (Fig.~\ref{fig:fig1}c). 
For both models, the allocation was relatively flat across depth, with exceptions near the beginning and end; attention taps were generally assigned lower resolution than MLP taps in the same layer. 
Consistently, the final readout was the most protected. 
As information was restricted further, a subset of taps reached the minimum realizable rate. 
In GPT-2, both layer 1 reads and the layer 11 attention read reached the rate floor first ($B=1.5\times10^4$), presumably preserving rate for more functionally important reads.

\begin{figure*}[h]
\vskip 0.2in
\begin{center}
\centerline{\includegraphics[width=\textwidth]{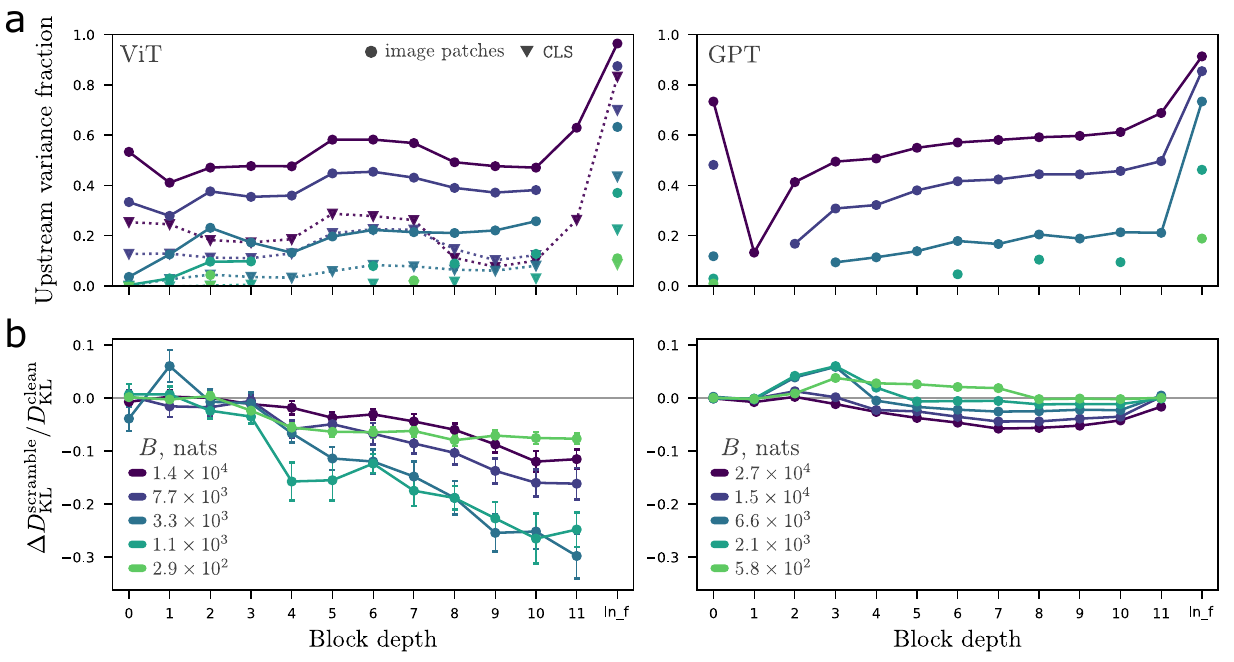}}
\caption{
\textbf{The nature of the compounded stochasticity.}
\textbf{(a)} The upstream variance estimated at each MLP tap, relative to the total (upstream and local posterior added variance), for ViT (left) and GPT2 (right) models.
For ViT, the \texttt{CLS} stream has different spread than the image patch representations.
The markers for collapsed taps in the lower information models are excluded.
\textbf{(b)} Change in distillation loss as a result of scrambling joint residual stream structure at layer $l$.
For multiple passes on the same input, the residual stream states are scrambled, breaking any joint structure in the effective posterior across streams.
The fractional change in the distillation loss $D_\text{KL}$ shows that scrambling improves the performance of ViT models and has a more muted effect on GPT models.
}
\label{fig:2_stochasticity_structure}
\end{center}
\vskip -0.2in
\end{figure*}

\subsection{Is the induced stochasticity structured?}
Noise introduced at each read propagates through subsequent blocks, producing compounding variation along the depth of the model. 
We first characterized this accumulation by decomposing the variance of repeated draws at each MLP tap into locally introduced and upstream-induced components (Fig.~\ref{fig:2_stochasticity_structure}a). 
In the ViT, patch streams accumulated substantially more upstream-induced variance than the \texttt{CLS} stream, whose upstream variation remained smaller than the noise introduced by its own reads. 
At the final tap, the unusually precise local read causes inherited variation to account for a correspondingly large fraction of the total variance.

We next probed dependencies across residual streams by independently shuffling each position across a batch of repeated draws. 
This approximately replaces samples from the joint distribution
$p(u_1,\ldots,u_T\mid\vec{x})$
with samples from the product of its position-wise marginals, while preserving both each marginal and the full input sequence $\vec{x}$. 
The resulting representation combines streams from different upstream noise trajectories and is therefore not an alternative available to the model within a single forward pass.

The effect was strongly model- and depth-dependent. 
In the ViT, shuffling reduced the distillation loss, and increasingly so at later depths.
In GPT-2, early shuffling produced small, budget-dependent degradations, whereas later shuffling was mildly beneficial.
One possible explanation lies in the interaction between attention structure and data modality. 
The ViT aggregates many partially redundant noisy measurements, for which removing cross-stream noise dependence can improve aggregation.
Causal language modeling instead builds directed, nested dependencies among non-redundant prefix representations, making coherent stochastic trajectories particularly important early in processing.

\begin{figure*}[h]
\vskip 0.2in
\begin{center}
\centerline{\includegraphics[width=\textwidth]{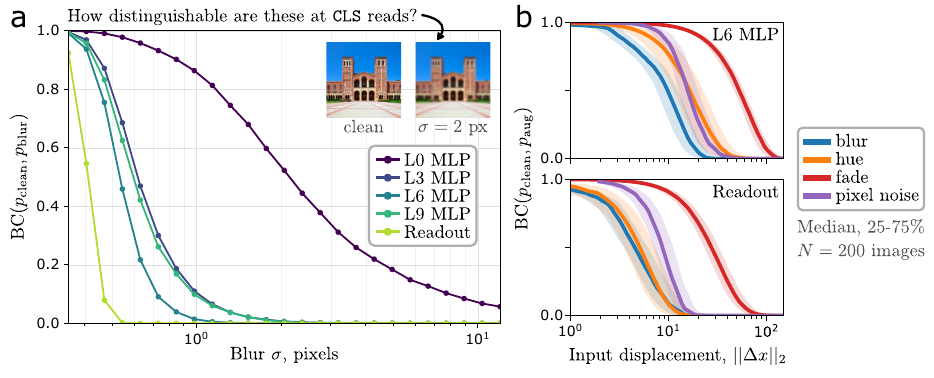}}
\caption{
\textbf{Distinguishability zone via continuous augmentations}
\textbf{(a)} For the ViT-S with $B=3.3 \times 10^3$ nats total budget, the BC of an image and its blurred version smoothly decreases---indicating increasing distinguishability---with the blur magnitude, and varies significantly across depth.
\textbf{(b)}
The model is more sensitive to some augmentations than others, shown here over 200 images via median curves with interquartile range shaded.}
\label{fig:3_visual_resolution}
\end{center}
\vskip -0.2in
\end{figure*}

\subsection{Vision resolution: Response curves to continuously graded augmentations}

In a ViT, the \texttt{CLS} stream is central to prediction yet identical for all images at initialization, making it a natural location to monitor where image distinctions emerge.
We applied continuously graded augmentations and measured the distinguishability of the resulting \texttt{CLS} reads from those of the unmodified image. 
In Fig.~\ref{fig:3_visual_resolution}a, progressively blurring an example image produces a corresponding decay in BC across depth.

For 200 images from the ImageNet-1k validation set, we applied four augmentations: blurring, hue rotation, additive Gaussian noise, and fading via contrast reduction. 
Indexing their magnitudes by the total Euclidean pixel displacement $||\Delta x||_2$ allows their effects to be compared at matched input-space distances. 
At both the layer 6 MLP and the final readout, the model was most sensitive to blurring and least sensitive to fading (Fig.~\ref{fig:3_visual_resolution}b). 
The relative ordering was broadly preserved at the final readout, while all augmentations became more distinguishable in absolute terms. 
These response curves trace augmentation-specific slices through a fuzzy resolution neighborhood around each image, relating graded changes in input space to when they become resolved during processing.

\begin{figure*}[h]
\vskip 0.2in
\begin{center}
\centerline{\includegraphics[width=\textwidth]{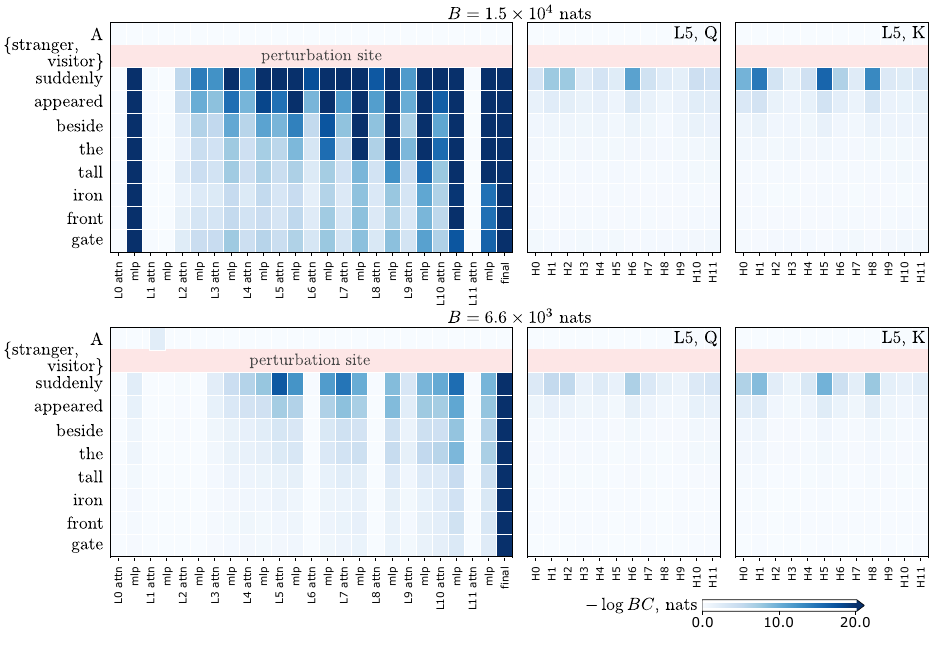}}
\caption{
\textbf{Perturbation distinguishability across the sequence and in specific heads.}
For the sentence shown at left, the perturbation consists of switching the \textit{stranger} token for \textit{visitor}.
Before the perturbation, in the residual stream of \textit{A}, there can be no distinguishability.  
After the perturbation, the distinguishability (here displayed by Bhattacharyya distance, or $- \log BC$) between streams depends on location in the sequence and in the processing.
Attention heads' projections tap into the distinguishability (\textit{right}); shown are the query (Q) and key (K) projections for the twelve heads of layer 5.}
\label{fig:4_text_perturbation_example}
\end{center}
\vskip -0.2in
\end{figure*}

\subsection{Natural language: Attention head selectivity}
In contrast to vision, textual perturbations are discrete in nature, and there is no single reference stream to compare across examples. 
To monitor the emergence of distinguishability, we perturbed a text sequence at a single location and left the remainder identical.
An example is shown in Fig.~\ref{fig:4_text_perturbation_example}, where the pair of sentences \texttt{A \{stranger/visitor\} suddenly appeared beside the tall iron front gate} differs only in the second token position.
The processing of both sentences cannot differ before the location of the perturbation, and the location of the perturbation is trivially distinguishable from the outset.
All subsequent tokens' streams initialize at identical embeddings, allowing us to monitor the distinguishability between the pairs as a function of depth in processing.
The block-level taps show a gradually increasing distinguishability (quantified via increasing Bhattacharyya distance, $\text{BD}=-\log \text{BC}$) with depth.
For the model with the larger information budget ($B=1.5 \times 10^4$ nats), the distinguishability is high early on in processing, whereas the other model processes the pair more similarly until the last layers.

In our single example in Fig.~\ref{fig:4_text_perturbation_example} (right), the Q and K projections in layer 5 display head-specific behavior in the positions immediately following the perturbation.
Because attention heads' projections are noninvertible to a lower dimensional space, information is lost in general, and the question becomes whether the particular distinction between the pairs is present in the subspace selected by each head's projections.
Given that language model heads have repeatedly been shown to perform specialized roles~\citep{olsson2022context,wang2023IOI}, it is plausible that their usage of distinguishing information from the residual stream might serve to differentiate and illuminate their role in processing.

\begin{figure*}[h]
\vskip 0.2in
\begin{center}
\centerline{\includegraphics[width=\textwidth]{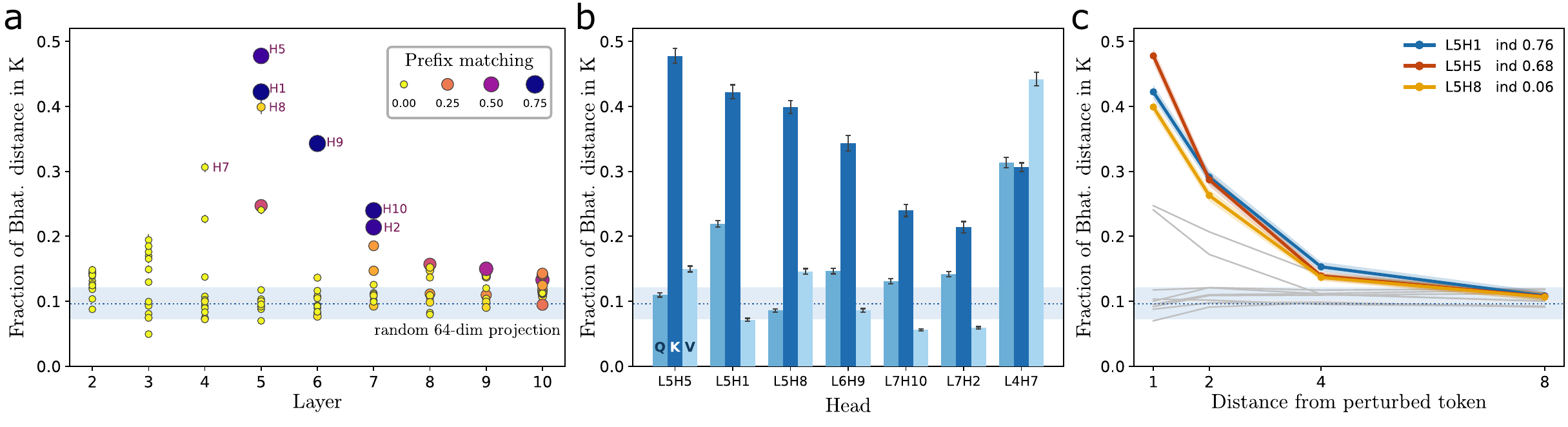}}
\caption{
\textbf{Key sensitivity to single-token perturbations.}
\textbf{(a)} The fraction of the Bhattacharyya distance preserved from each attention read into its subsequent K projections is shown for all heads between layers 2 and 10, for the $B=1.5 \times 10^4$ nats GPT-2 model.
For reference, random 64-dimensional projections preserve around 0.1 of the upstream Bhattacharyya distance (shaded band is 5-95\% spread).
Heads' key projections are colored and sized according to the head's prefix-matching score~\citep{olsson2022context}.
\textbf{(b)} The same fraction for the top heads in \textbf{a}, evaluated at the query (Q) and value (V) projections as well.
\textbf{(c)} The same fraction for all heads in layer 5 as a function of distance from the perturbation.
}
\label{fig:5_head_selectivity}
\end{center}
\vskip -0.2in
\end{figure*}

We generated 80 minimally perturbed pairs by replacing nouns in OpenWebText samples with randomly selected nouns from the same category. 
We retained swaps producing at most 50 nats of Bhattacharyya distance at the layer 5 attention read immediately following the perturbation. This controls perturbation severity: larger changes produce widespread distinguishability and obscure head-to-head selectivity.

Figure~\ref{fig:5_head_selectivity} shows the fraction of upstream distinguishability retained by each K projection in the GPT-2 model operating at $B=1.5 \times 10^4$ nats. Random 64-dimensional projections preserve a nonzero fraction and provide a baseline near which most learned K projections operate.

Seven heads in the middle layers stand out above this baseline. 
Five coincide with high-induction-score heads in the GPT-2 Small induction mosaic of \citet{nandaInductionMosaic}. Among the remaining two, \texttt{L5H8} was described as a ``fuzzy induction head'' in the canonical indirect object identification analysis; \texttt{L5H5} and \texttt{L6H9} were identified there as canonical induction heads \citep{wang2023IOI}. 
Thus, six of the seven K-selective heads have an independently observed connection to induction.

The correspondence is also reflected, though incompletely, in prefix-matching scores computed for the stochastic model from attention maps on repeated random-token sequences \citep{olsson2022context}. 
Attention maps indicate which positions a head weights, whereas K distinguishability identifies which perturbation-induced distinctions are exposed to its attention routing.
Notably, \texttt{L5H8} and \texttt{L4H7} retain strong K selectivity despite having nearly zero prefix-matching scores. 
The former's previous characterization as a fuzzy induction head illustrates how distinction preservation can reveal induction-adjacent structure that is weak or absent in the canonical attention statistic. 
For all highlighted heads except \texttt{L4H7}, this selectivity is specific to K: the corresponding Q and V projections remain near the random-projection baseline (Fig.~\ref{fig:5_head_selectivity}b). 
\texttt{L4H7} instead preserves unusually high distinguishability across Q, K, and V, suggesting broader sensitivity to the perturbation. 
Finally, for the layer 5 heads, excess K selectivity decays with distance from the perturbed token toward the network-wide baseline (Fig.~\ref{fig:5_head_selectivity}c).

\section{Discussion}

For unconstrained point representations processed by an expressive function class, geometric proximity alone provides no general guarantee of similar downstream behavior. We introduce stochasticity during optimization so that transformer blocks instead learn to operate on overlapping distributions of residual-stream reads. 
This gives distinguishability an operational meaning: overlap bounds the distinctions transmitted through that read, although an attention block may still access them through other streams. 
The resulting analysis is nevertheless deliberately one-sided. High overlap constrains two inputs to be processed similarly, whereas low overlap says only that a distinction is available; it does not specify what differs or how the sublayer will use that distinction.

The construction should not be taken to imply that the point representations of an unmodified transformer possess an intrinsic probabilistic geometry. 
Rather, it produces a controlled, finite-precision extension of the pretrained model: with few added parameters and modest fine-tuning, the model can remain within hundredths of a nat of the original behavior while every residual-stream read is made stochastic and rate-limited. 
The lower-rate boundary of this behaviorally anchored regime thereby offers a probe of the original model’s operating resolution.

Several approximations remain. The information-rate expression is an upper bound whose tightness depends on the distribution of measured states over the LayerNorm sphere. Moreover, our tractable vMF analysis describes local posteriors, while upstream stochasticity induces effective posteriors that are generally mixtures and may be substantially broader. Common random numbers allow us to approach these effective posteriors incrementally and to compare paired computations with reduced variance, but they do not eliminate the difficulty of characterizing the full mixture.

Our experiments on ViT-S and GPT-2 small should therefore be viewed as a proof of concept. Further work should test whether the observed patterns persist at larger scales, develop better approximations to effective posteriors, and extend the analysis from individual reads and projections to the joint distinctions available to an attention block. It will also be important to determine when conclusions obtained from rate-limited models transfer to their unconstrained counterparts. More broadly, the framework reframes representational analysis from asking which internal states are geometrically similar to asking which input distinctions remain available, where they are discarded or introduced, and which components are able to act on them. This perspective is intentionally narrower than a complete account of computation, but it provides a functional foundation on which more detailed mechanistic analyses can build.

\bibliography{references}
\bibliographystyle{iclr2027_conference}

\clearpage

\appendix
\section{Implementation}
\label{appx:implementation}

Code for this project can be found at \href{https://murphyka.github.io/stoch_layernorm}{murphyka.github.io/stoch\_layernorm}.  
Excepting the analysis, the crux is the simple modification to LayerNorm, presented in Alg.~\ref{alg:noisy_layernorm}.

\begin{algorithm}[t]
\caption{Noisy LayerNorm Read}
\label{alg:noisy-ln}
\begin{algorithmic}[1]

\Statex
\Statex \textbf{(Replaces every \texttt{nn.LayerNorm}).}
\Function{NoisyLN}{$x;\ \gamma, \beta, \sigma$}
  \State $\hat{x} \gets \Normalize(x)$, \quad $\Normalize(u) := \dfrac{u - \mathrm{mean}(u)}{\sqrt{\mathrm{var}(u) + \epsilon}}$
    \Comment{statistics over the width axis}
  \State $z \sim \mathcal{N}(0, I_D)$
    \Comment{drawn independently per token/patch position and per sample}
  \State $\hat{x} \gets \Normalize(\hat{x} + \sigma z)$
    \Comment{re-normalise: the read stays on $\mathcal{S}^{D-2}$}
  \State \Return $\gamma \odot \hat{x} + \beta$
    \Comment{learned affine applied \emph{after} the channel}
\EndFunction

\end{algorithmic}
\label{alg:noisy_layernorm}
\end{algorithm}

\subsection{The channel and its calibration}
\label{app:channel}

Every \texttt{nn.LayerNorm} in the network is replaced by a noisy read: the input is normalised,
isotropic Gaussian noise of scale $\sigma$ is added, the result is re-normalised, and only then are
the learned gain and bias applied. Because a LayerNorm output has zero mean and unit RMS, each read
lives on the sphere $\mathcal{S}^{D-2}$ inside the mean-zero hyperplane of $\mathbb{R}^{D}$. 
We approximate each read as $\mathrm{vMF}(\mu, \kappa)$ and take the
concentration $\kappa$ as the rate knob:
\begin{equation}
  \mathrm{rate}(\kappa)
  = \mathrm{KL}\!\left(\mathrm{vMF}(\mu,\kappa) \,\|\, \mathrm{Unif}\right)
  = \kappa\,A_p(\kappa) + \log C_p(\kappa) - \log C_p(0),
  \label{eq:rate}
\end{equation}
in closed form, and the sampling scale is matched to it through the mean resultant length
$\rho = A_p(\kappa) = I_{p/2}(\kappa) / I_{p/2-1}(\kappa)$ by $\sigma = \sqrt{1/\rho^{2} - 1}$, so
that the mean read-cosine of the add-Gaussian-then-renormalise channel equals $\rho$ (verified
numerically against the analytic $\rho$, and the closed-form rate against a Wood/Ulrich Monte-Carlo
estimate). 
This matching is what makes the channel reparameterisation-clean: gradients flow to both
the weights and the rate allocation with no rejection sampler in the loop. Training uses a
precomputed strictly-monotone $4000$-point geometric grid in $\kappa$ (up to $10^{6}$) with the rate
and $\sigma$ columns exposed as differentiable one-dimensional linear interpolants.

Both ends of that grid are set by floating-point limits. 
All Bessel evaluations use the
exponentially scaled modified Bessel function of the first kind,
$\widetilde{I}_v(z) = I_v(z)\,e^{-z}$ for real $z > 0$ (SciPy's \texttt{scipy.special.ive}): the
unscaled $I_v$ overflows to infinity at the top of the grid for every $p$ we use, whereas the scaling
cancels in the ratio $A_p$ and enters the log-normaliser additively as
$\log I_v(\kappa) = \log \widetilde{I}_v(\kappa) + \kappa$. 
At the bottom of the grid the opposite
failure appears, and it is \emph{not} an artefact of the scaling: in double precision
$I_{p/2-1}(\kappa)$ underflows to exactly zero for $\kappa \lesssim 20$ at $p = 767$, scaled or not,
which makes $\rho = A_p(\kappa)$ a $0/0$. The ViT runs therefore keep the default floor
$\kappa_{\min} = 5$, which stays representable for $p \le 383$, while for GPT-2 the floor is found by
bisection as the smallest $\kappa$ whose $\widetilde{I}_v$ clears a conservative $10^{-250}$, giving
$\kappa_{\min} = 76.3$ at $p = 767$. Since the interpolants clamp their argument to the grid, this
floor also caps the largest representable per-tap noise: $\sigma \le 76.6$ at
$p = 383$ and $10.1$ at $p = 767$. The cap is reached in practice, as the learned allocation drives
the most heavily compressed taps down onto it: for GPT-2, $5/25$ taps sit at the cap at
$\sigma_g = 1$, rising to $13/25$ at $\sigma_g = 2$ and $22/25$ at $\sigma_g = 8$ (ViT-Small,
$1/25$ and $18/25$ at $\sigma_g = 1$ and $8$).

\subsection{Rate budget and allocation}
\label{app:budget}

A run is specified by a single dimension-free knob, the uniform-equivalent noise level $\sigma_g$.
This is converted to $\kappa_g$, then to a per-tap rate, and the \emph{total} budget is fixed at
$B = n \cdot \mathrm{rate}(\kappa_g)$ over the $n$ taps; the network learns only how to spend it, via
unconstrained logits $a$ passed through a softmax,
\begin{equation}
  r_i = B \cdot \mathrm{softmax}(a)_i,
  \qquad
  \textstyle\sum_{i=1}^{n} r_i = B \ \ \text{by construction}.
  \label{eq:alloc}
\end{equation}
Conserving the budget in \emph{rate} (rather than penalising it) removes the degenerate solution
where the model compresses every tap and pays the penalty once, and no penalty term appears in the
loss at all: the distillation KL alone drives a water-filling allocation. 
Both architectures contain 25 taps, although the rate associated with a fixed \(\sigma_g\) depends on the representation dimension.

\subsection{The noise ramp}
\label{app:ramp}

We ramp the noise budget: the run starts at a near-clean $\sigma_g = 0.05$ and
interpolates to the target over the first $6$ of $25$ epochs ($\sim$24\% of training), updated
\textbf{every optimiser step} rather than per epoch. 
Two interpolation shapes are
implemented: linear in $\log$-rate and linear in $\log \sigma_g$. 
The GPT-2 sweep used the former, and ViT the latter.
Everything else is shared: AdamW ($\beta_1 = 0.9$, $\beta_2 = 0.999$) with two
parameter groups --- network weights at the run's learning rate with weight decay $5\times10^{-3}$,
allocation logits at $6\times10^{-3}$ with \textbf{no} decay --- cosine decay after a $0.2$-epoch
linear warmup, global grad-norm clipping at $1.0$ over weights and logits jointly, \texttt{bfloat16}
autocast on A100 MIG partitions. 
Every run is a \emph{finetune} arm
whose loss is $\mathrm{KL}(\text{teacher} \,\|\, \text{noisy student})$ at $T = 1$ against a frozen
copy of the pretrained weights, with the student initialised at the teacher; distilling from the
clean model rather than from labels keeps the distortion axis measured in the same units across both
modalities.

\subsection{GPT-2}
\label{app:gpt2}

GPT-2-small ($124$M, $D = 768$; \texttt{ln\_1}/\texttt{ln\_2} per block plus \texttt{ln\_f}, giving
$25$ taps) is finetuned on OpenWebText~\citep{Gokaslan2019OpenWeb}.
Documents get an explicit EOS appended before packing, so GPT-2's own
context-reset convention sits at every document boundary instead of unrelated documents being
spliced together; blocks are non-overlapping $512$-token windows, which removes any need for padding
or attention masks. 
Validation is a held-out contiguous $5000$-document tail of the same stream. 
Runs use batch $32 \times 512$ tokens with \texttt{limit\_train\_batches} $= 5000$, i.e.\ $81.9$M tokens
per epoch and $\approx 2.05$B tokens over $25$ epochs --- under a quarter of one pass, so no example
is seen twice. 
Learning rate $1\times10^{-4} \to 1\times10^{-6}$ cosine. 
All three dropout probabilities (residual, embedding, attention) are forced to $0.0$: dropout is an uncontrolled second noise source that would sit underneath the vMF channel during training and confound the rate measurement. 
The loss is a manually shifted cross-entropy, and the distillation KL is chunked in
blocks of $128$ positions along the sequence --- the full $[B, L, V]$ \texttt{float32} softmax at
vocabulary size $50257$ is $\sim$3.3\,GB per tensor, enough to decide whether a run fits its MIG
slice --- with identical \texttt{float32} arithmetic, not a precision trade. 
Evaluation is full-validation perplexity plus next-token top-1 accuracy under the noise (never a clean or mean
pass). 
Note that a GPT-2 ``epoch'' here is a fixed $5000$ optimiser steps, not a pass over the corpus.

\subsection{ViT}
\label{app:vit}

The vision runs finetune timm's 
\texttt{vit\_small\_patch16\_224} ($D = 384$) on full ImageNet-1k at $224^{2}$. 
Preprocessing is timm's standard ViT config ---
bicubic interpolation, \texttt{crop\_pct} $0.9$, mean/std $0.5$ --- with deliberately
\emph{finetuning}-weight augmentation held fixed across the whole sweep so that cross-$\sigma_g$ and
cross-model comparisons are not confounded by an augmentation change: RandAugment
\texttt{rand-m7-mstd0.5-inc1}, RandomResizedCrop with the scale floor raised to $0.4$ (no extreme
micro-crops), horizontal flip $0.5$, no random erasing, and drop-path $0$. 
Mixup/CutMix are disabled
in the distillation arm --- the loss is the teacher KL, so soft-target cross-entropy and label
smoothing do not apply. 
Batch $512$, learning rate $4\times10^{-4} \to 1\times10^{-5}$ cosine, $25$
epochs (a $50$-epoch arm was also run and did not change the picture). 
Validation accuracy is
computed on a fixed $40$-batch ($20{,}480$-image) subsample of the held-out split, with the channel
\emph{on} at the run's learned allocation.
An ImageNet epoch here is one full pass over the training set.

\section{Gaussian to von Mises-Fisher approximation}
\label{appx:gaussian_vmf_calibration}

\begin{figure*}[h]
\vskip 0.2in
\begin{center}
\centerline{\includegraphics[width=\textwidth]{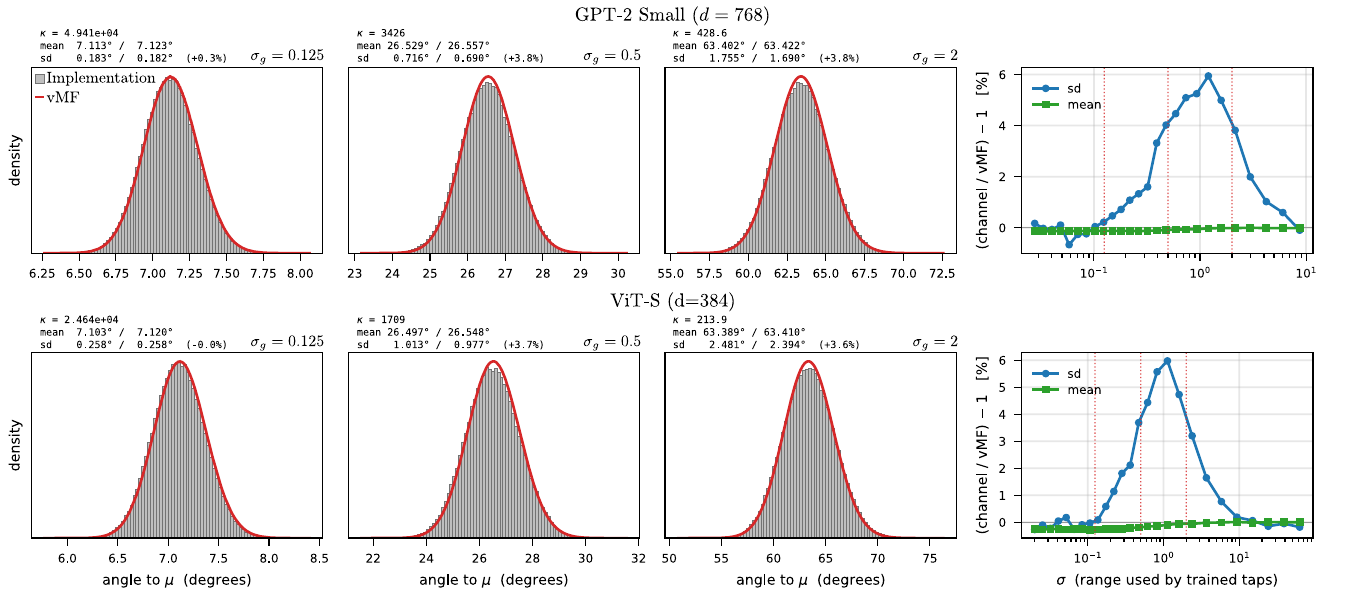}}
\caption{
\textbf{Calibration of the von Mises-Fisher fit to the noisy LayerNorm channel.}
For several levels of noise, we show the histogram of angular deviations from the mean, $\mu$ (gray), and the analytic vMF density (red) used in our rate and Bhattacharyya coefficient calculations.
The two means and spreads are printed above the three examples at the GPT-2 (top) and ViT-S (bottom) dimensionalities.
On the right, the difference in the mean and spread of angular deviations is measured across a range of noise levels.
}
\label{fig:vmf_calibration}
\end{center}
\vskip -0.2in
\end{figure*}

In Fig.~\ref{fig:vmf_calibration}, we compare the angular spread induced by the noisy LayerNorm operation to that of an approximating vMF distribution.
Due to rotational symmetry, the comparison reduces to a single scalar: the angular deviation from the distribution mean, $\mu$.

\section{Pushforward density under query, key, and value projections}
\label{appx:qkv_pushforward}

Let
\[
    \mathcal{H}
    =
    \left\{
        z\in\mathbb{R}^d:
        \mathbf{1}^{\top}z=0
    \right\}
\]
denote the mean-zero LayerNorm subspace, and let $p=d-1$ be its dimension. Choose an orthonormal basis $E\in\mathbb{R}^{d\times p}$ for $\mathcal{H}$. Under the unit-radius convention used for the vMF approximation, a stochastic read $V\in\mathcal{H}$ can be written as $V=EX$ for some $X\in\mathcal{S}^{p-1}$. The corresponding unscaled LayerNorm state is $\sqrt{d}EX$.

For a query, key, or value projection $A$, including the LayerNorm affine parameters $\gamma$ and $\beta$, the projected state is
\[
    Y_A
    =
    A\left(\sqrt{d}\,\gamma\odot EX+\beta\right)
    =
    B_AX+b_A,
\]
where
\[
    B_A
    =
    \sqrt{d}\,
    A\operatorname{diag}(\gamma)E,
    \qquad
    b_A=A\beta.
\]
The shift $b_A$ is common to all local posteriors and does not affect their BC.

If $B_A$ is rank deficient, we restrict its codomain to $\operatorname{im}(B_A)$ and express the output in orthonormal coordinates on this image. The resulting matrix has full row rank; we continue to denote it by $B_A$ and let $m$ denote its output dimension. Define
\[
    S_A=(B_AB_A^\top)^{1/2},
    \qquad
    P_A=S_A^{-1}B_A.
\]
Then
\[
    B_A=S_AP_A,
    \qquad
    P_AP_A^\top=I_m.
\]
Thus, $P_A$ has orthonormal rows and $S_A$ is invertible. Writing $W_A=P_AX$, we have $Y_A-b_A=S_AW_A$. Translation by $b_A$ and transformation by $S_A$ are both invertible on the support of the projected distributions, so
\[
    \operatorname{BC}
    \left(
        p_{Y_A\mid u_i},
        p_{Y_A\mid u_j}
    \right)
    =
    \operatorname{BC}
    \left(
        p_{W_A\mid u_i},
        p_{W_A\mid u_j}
    \right).
\]
Because $P_A$ has orthonormal rows and $X$ has unit norm, the support of $W_A$ is the unit ball
\[
    \mathbb{B}^m
    =
    \left\{
        w\in\mathbb{R}^m:
        \lVert w\rVert_2\leq 1
    \right\}.
\]

We next derive the density on this ball. At a fixed tap, consider a local posterior
\[
    X\sim\operatorname{vMF}_p(\mu,\kappa),
    \qquad
    p(x\mid\mu,\kappa)
    =
    C_p(\kappa)\exp(\kappa\mu^\top x).
\]
Let $q=p-m$. Complete the rows of $P_A$ with an orthonormal basis $D\in\mathbb{R}^{q\times p}$ for its null space, so that the matrix formed by stacking $P_A$ and $D$ is orthogonal. Every point in the spherical preimage of $w\in\mathbb{B}^m$ can then be written as
\[
    x
    =
    P_A^\top w
    +
    D^\top
    \sqrt{1-\lVert w\rVert_2^2}\,\xi,
    \qquad
    \xi\in\mathcal{S}^{q-1}.
\]
The spherical surface measure decomposes as
\[
    d\omega_{p-1}(x)
    =
    \left(1-\lVert w\rVert_2^2\right)^{q/2-1}
    dw\,d\omega_{q-1}(\xi).
\]

Define
\[
    a=P_A\mu,
    \qquad
    \rho=\lVert D\mu\rVert_2
    =
    \sqrt{1-\lVert a\rVert_2^2}.
\]
For a point in the preimage of $w$,
\[
    \mu^\top x
    =
    a^\top w
    +
    \sqrt{1-\lVert w\rVert_2^2}\,
    (D\mu)^\top\xi.
\]
Integrating over $\xi$ and using the vMF normalizing constant in dimension $q$ gives the pushforward density
\[
    g_\mu^A(w)
    =
    \frac{
        C_p(\kappa)
    }{
        C_q\left(
            \kappa\rho
            \sqrt{1-\lVert w\rVert_2^2}
        \right)
    }
    \exp(\kappa a^\top w)
    \left(1-\lVert w\rVert_2^2\right)^{q/2-1},
    \qquad
    w\in\mathbb{B}^m.
\]

For local posteriors centered at $\mu_i$ and $\mu_j$, let $g_i^A$ and $g_j^A$ denote their respective pushforward densities. Their projected BC is
\[
    \operatorname{BC}_A(u_i,u_j)
    =
    \int_{\mathbb{B}^m}
    \sqrt{
        g_i^A(w)g_j^A(w)
    }\,dw.
\]
We estimate this integral using the equally weighted proposal
\[
    r_A(w)
    =
    \frac{1}{2}
    \left[
        g_i^A(w)+g_j^A(w)
    \right].
\]
Samples from $r_A$ are obtained by selecting $i$ or $j$ with equal probability, drawing $X$ from the selected vMF local posterior, and computing $W_A=P_AX$. Importance sampling then gives
\[
    \widehat{\operatorname{BC}}_A(u_i,u_j)
    =
    \frac{1}{N}
    \sum_{n=1}^{N}
    \frac{
        2\sqrt{
            g_i^A(w_n)g_j^A(w_n)
        }
    }{
        g_i^A(w_n)+g_j^A(w_n)
    },
    \qquad
    w_n\sim r_A.
\]
This estimator is unbiased under exact sampling. Moreover, the arithmetic--geometric mean inequality gives
\[
    0
    \leq
    \frac{
        2\sqrt{
            g_i^A(w)g_j^A(w)
        }
    }{
        g_i^A(w)+g_j^A(w)
    }
    \leq
    1,
\]
so every Monte Carlo summand is bounded.

\end{document}